\documentclass{article}
\usepackage{spconf,amsmath,epsfig}
\usepackage{graphicx}
\usepackage{subcaption}
\usepackage{amsmath}
\usepackage{algorithm}
\usepackage{algorithmic}
\usepackage{booktabs}
\usepackage{float}
\usepackage{yhmath}
\usepackage[backref, colorlinks=true,citecolor=green,urlcolor=pink,linkcolor=red]{hyperref}
\usepackage[doipre={doi:~}]{uri}
\usepackage{bm}
\usepackage{url}

\let\OLDthebibliography\thebibliography
\renewcommand\thebibliography[1]{
  \OLDthebibliography{#1}
  \setlength{\parskip}{0pt}
  \setlength{\itemsep}{0pt plus 0.3ex}
}

\begin{document}
\title{Unsupervised Severely Deformed Mesh Reconstruction (DMR) from a Single-View Image}
\name{\normalsize Jie Mei$^{\star}$\thanks{$^{\star}$e-mail:\{jiemei, jingxiyu, hwang\}@uw.edu}, Jingxi Yu$^{\star}$, Suzanne Romain$^{\dagger}$, Craig Rose$^{\dagger}$, Kelsey Magrane$^{\dagger}$\thanks{$^{\dagger}$e-mail:\{suzanne.romain, craig.rose, kelsey.magrane, graeme.leeson\}\@noaa.gov}, Graeme LeeSon$^{\dagger}$,  Jenq-Neng Hwang$^{\star}$}

\address{$^{\star}$University of Washington, Seattle, WA, USA\\ $^{\dagger}$Pacific States Marine Fisheries Commission, National Oceanic and Atmospheric Administration, USA}

\maketitle
\begin{abstract}
Much progress has been made in the supervised learning of 3D reconstruction of rigid objects from multi-view images or a video. However, it is more challenging to reconstruct severely deformed objects from a single-view RGB image in an unsupervised manner. Although training-based methods, such as specific category-level training, have been shown to successfully reconstruct rigid objects and slightly deformed objects like birds from a single-view image, they cannot effectively handle severely deformed objects and neither can be applied to some downstream tasks in the real world due to the inconsistent semantic meaning of vertices, which are crucial in defining the adopted 3D templates of objects to be reconstructed. In this work, we introduce a template-based method to infer 3D shapes from a single-view image and apply the reconstructed mesh to a downstream task, i.e., absolute length measurement. Without using 3D ground truth, our method faithfully reconstructs 3D meshes and achieves state-of-the-art accuracy in a length measurement task on a severely deformed fish dataset.
\end{abstract}
\section{Introduction}
\label{sec:intro}
Modeling the 3D geometry of objects is an open research problem in computer vision. 3D shape reconstruction enables machines to perceive the 3D world. Given 2D RGB inputs, predicting the 3D shape of objects looks natural to humans but quite challenging to machines. The key problem is the ill-posed nature of the task, i.e., with limited 2D observed views, the object's shape can be arbitrary in any unseen view.

To deal with this ill-posed problem, a promising approach is to use supervised learning approaches requiring 3D ground truth, such as Mesh R-CNN~\cite{Gkioxari_2019_ICCV} and Pixel2Mesh~\cite{wang2018pixel2mesh}. Although supervised training has been quite successful in 2D tasks such as image classification~\cite{ mei2021videobased}, object detection~\cite{ren2016faster} and segmentation~\cite{ronneberger2015u}, it is much more time-consuming and expensive to collect 3D ground truth data for 3D supervised training. For example, SMPL~\cite{loper2015smpl} uses thousands of registered 3D scans of humans for training, SMAL~\cite{zuffi20173d} utilizes scans of animal toys and a manually rigged mesh model. Collecting such 3D ground truth data requires much more time and effort than collecting 2D data. Therefore, existing supervised training methods fail to reconstruct objects' shapes beyond the training categories, particularly for objects whose 3D ground truths are impossible to collect such as wild animals.

Without supervised training, a recent trend to mitigate this ill-posed problem is leveraging category-specific image collections for unsupervised training. By using images from the same category for training, existing methods, such as CMR ~\cite{kanazawa2018learning} and ACSM~\cite{kulkarni2020articulationaware}, assume the model can learn the common shape structure in the category. However, such results are limited to rigid objects or slightly deformed objects such as birds. Besides, these methods can not be applied to some downstream tasks, such as size measurement, because they can not guarantee the consistent semantic meaning of each vertex in the 3D mesh template. 

We proposed an unsupervised learning method, which lies in an intermediate regime between the above two extremes, keeping the model's ability to reconstruct severely deformed objects and at the same time maintaining the semantic meaning of each mesh vertex. Our proposed method can thus ensure the reconstructed shape to be applied to some downstream tasks, that require consistent semantic meaning of mesh vertices, such as length measurement.

\textbf{Why single-view?} 3D reconstruction models, such as Nerf~\cite{mildenhall2020nerf}, uses multi-view images as input while CMR~\cite{kanazawa2018learning}, ACSM~\cite{kulkarni2020articulationaware} and LASR~\cite{yang2021lasr} use a single view image. Though achieving better performance, multi-view data are commonly much harder to collect than single-view images or monocular videos. Previous methods~\cite{huang2018fish, mei2021absolute} also mention the challenges encountered in the synchronization of multi-view cameras. Therefore, our proposed method is based on a single-view image.

\textbf{Why a template?} In order to maintain the consistent semantic meaning of each mesh vertex defined for 3D objects, prior works introduce some templates, such as the 3D human template in SMPL~\cite{loper2015smpl}. Because only when each mesh vertex has consistent semantic meaning, the reconstructed shape can be further applied in some downstream tasks. Our method also adopts a 3D template for unsupervised estimation of the 3D object poses under severely deformed scenarios.

\begin{figure*}[htpb]
\centering
\includegraphics[width=1\textwidth]{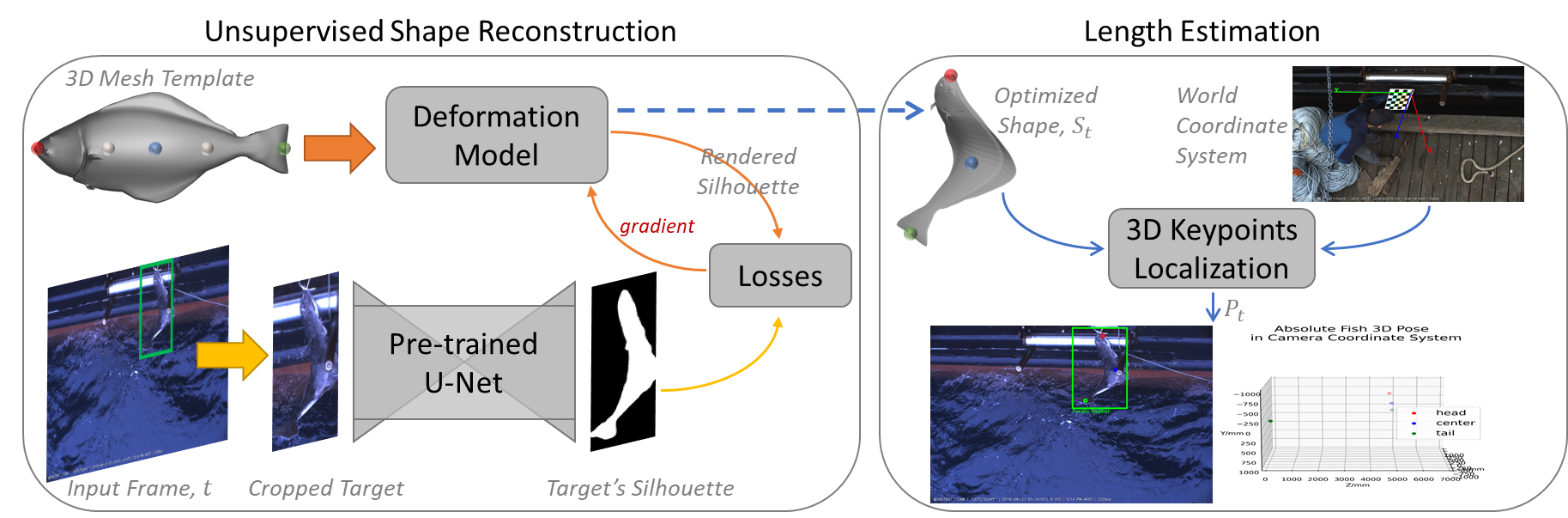}
\caption{Pipeline overview: Given a single input frame with the segmentation mask of the target fish, we reconstruct the deformed pose and 3D shape \bm{$S_t$} and measure its length by localizing the absolute 3D keypoints \bm{$P_t$} in the camera coordinate system. The optimized shape \bm{$S_t$} is represented by a 3D mesh template with deformation parameters under linear blend skinning (Sec.~\ref{sec:def}). The unsupervised shape reconstruction is a gradient-based optimization (Sec.~\ref{sec:unsup}) and the length estimation is based on our 3D localization method (Sec.~\ref{sec:Loc}).  *Note: In the world coordinate system image, the red axis is $Z$ axis. In the camera coordinate system image, the camera image plane is on $Z=0$ plane, and $Z$ axis is depth, with the unit being millimeter. So, the demo fish is about 5 meters away from the camera. Three pre-defined keypoints are head, center, and tail points shown in red, blue, and green dots respectively. Two pearl dots shown in the 3D mesh template are pre-defined joints inside the fish body.}
\label{pipeline}
\end{figure*}

\textbf{Proposed approach:} Instead of predicting deformed poses and 3D shapes from category-specific image collections or multi-view inputs, we propose a shape model with the ability to severely deform for matching with a single view image of an object. We utilize a differentiable render~\cite{liu2019softras} to make whole learning unsupervised, given an initialized 3D template that can be deformed based on some learnable parameters. By comparing the forward rendered silhouette (segmentation mask) under these learnable parameters with the ground truth of object silhouette, we use Adam optimizer to optimize these deformation parameters so that we can reconstruct the object shape in an unsupervised manner without any 3D ground truth. An overview of the pipeline is shown in Fig.\ref{pipeline}

\textbf{Contributions} We propose an unsupervised learning method for severely deformed shape reconstruction based on a 3D deformable template. With the consistent semantic meaning of each mesh vertex, we are able to apply the reconstructed mesh to a downstream task, length measurement. By only using a target's 2D segmentation mask, our model can reconstruct the severely deformed shape and further estimate the 3D pose of the object, and perform the length measurement task from a monocular video, which further allows averaging estimated lengths for all frames corresponding to one object.

\section{Proposed Method}
\textbf{Problem} Given a monocular video containing an object of interest, i.e., severely deformed fish (Pacific Halibut) in the dataset, with ground truth of segmentation mask of the object, we tackle the severely deformed 3D shape reconstruction and length estimation problem, which includes estimating (1) \bm{$S_t:$} the relative 3D shape of the target for each frame $t$, (2) \bm{$P_t:$} the absolute 3D keypoints' locations of the target in the camera coordinate system for each frame $t$.

\textbf{Overview} As shown in Fig~\ref{pipeline}, the whole pipeline includes two parts: unsupervised shape reconstruction and length estimation. The deformation model in the unsupervised shape reconstruction part contains learnable parameters to severely deform the 3D mesh template (Sec.~\ref{sec:def}). Inspired by recent works in differentiable rendering~\cite{liu2019softras, liu2020general,ravi2020pytorch3d}, our unsupervised shape reconstruction part also uses a differentiable render to generate the silhouette of the severely deformed 3D mesh. With the ground-truth detection bounding boxes in the dataset, we first crop the target fish and then use a pre-trained U-Net~\cite{ronneberger2015u}, serving as an instance segmentation mechanism, to obtain the pseudo ground-truth silhouette. Deformation model parameters \bm{$W$} are then updated to minimize the difference between the rendered silhouette and  the pseudo ground-truth (Sec.~\ref{sec:unsup}). Following MonoPose~\cite{mei2021absolute}, our length estimation part also introduces a world coordinate system to transfer the relative optimized shape \bm{$S_t$} to the absolute location in the camera coordinate system (Sec.~\ref{sec:Loc}).

\subsection{Deformation Model}
\label{sec:def}
Inspired by template-based methods~\cite{wang2019multi, yang2021lasr, mei2021absolute, loper2015smpl}, we introduce a relative 3D mesh fish template. In order to deform this template, we pre-define two joints inside the template body as shown with two pearl dots in the top left corner image in Fig~\ref{pipeline}. In our deformation model, we adopt linear-blend skinning (LBS) model~\cite{lewis2000pose} to deform our template 3D mesh.

\textbf{Linear-Blend Skinning (LBS)} uses weighted joints' deformation to deform each mesh vertex so that the deformation of the whole body is smooth and natural. Joints' deformation includes rotation, translation, and scaling denoted as $R_{j,t}$, $T_{j,t}$, $S_{j,t}$ respectively:

\begin{equation}
\begin{aligned}
Rotation: {V_{i,t}} &\Leftarrow  R_{0,t}\cdot(\sum_j{W_{j,i} \cdot R_{j,t}})\cdot V_{i} \\ Translation: {V_{i,t}} &\Leftarrow  T_{0,t}+(\sum_j{W_{j,i} \cdot T_{j,t}}) + V_{i},\\ Scaling: {V_{i,t}} &\Leftarrow  S_{0,t}\cdot(\sum_j{W_{j,i}\cdot S_{j,t}})\cdot V_{i},
\end{aligned}
\end{equation}
where $i$ is the 3D mesh vertex index, $t$ is the frame index, $j\in[1,2]$ is the joint index. $V_i\in R^3$ is the original vertex coordinate on the 3D mesh template. $V_{i,t}\in R^3$ is the vertex coordinate after deformation. $W_{j,i}$ is the skinning weighting on joint $j$'s deformation for vertex $i$. $R_{0,t}$, $T_{0,t}$, and $S_{0,t}$ are the deformation for the object root body, i.e., whole body.

Although the same fish species share a similar body shape, the individual fish shapes can be slightly different from each other. Rotation of joints and root body can transform the template to coarsely match with most severely deformed shapes, while the use of translation and scaling deformation is to further model the fine-grained difference between the shared template and individual fish. 

\textbf{Learnable Skinning Weights} We model the skinning weights $W_{j,i}$ as a mixture of Gaussians, which is inspired by the local 3D shape learning works~\cite{sumner2007embedded, genova2020local}, where two components are used because our template has two joints:

\begin{equation}
\begin{aligned}
{W_{j,i}} &=  C \cdot e^{-1/2 (V_i - V_j)^T Q_{j,t} (V_i-V_j)},
\end{aligned}
\end{equation}
where $C$ is a normalization factor to make sure the weightings on all joints for one mesh vertex sum up to one. $V_i\in R^3$ is the original vertex coordinate on the 3D template. $V_j \in R^3$ is the original coordinate of joint $j$ (pearl dots). $Q_{j,t}\in R^{3\times3}$ is the covariance matrix that determines the orientation and radius of the Gaussian. Obviously, this weighting is based on the distance between each mesh vertex and each joint. If a mesh vertex is closer to a joint, that joint's deformation will contribute more to that vertex's deformation. Since $Q_{j,t}$ is a learnable matrix, the corresponding weighting $W_{j,i}$ is thus learnable.

\begin{figure}[htpb]
\centering
\includegraphics[width=0.52\textwidth]{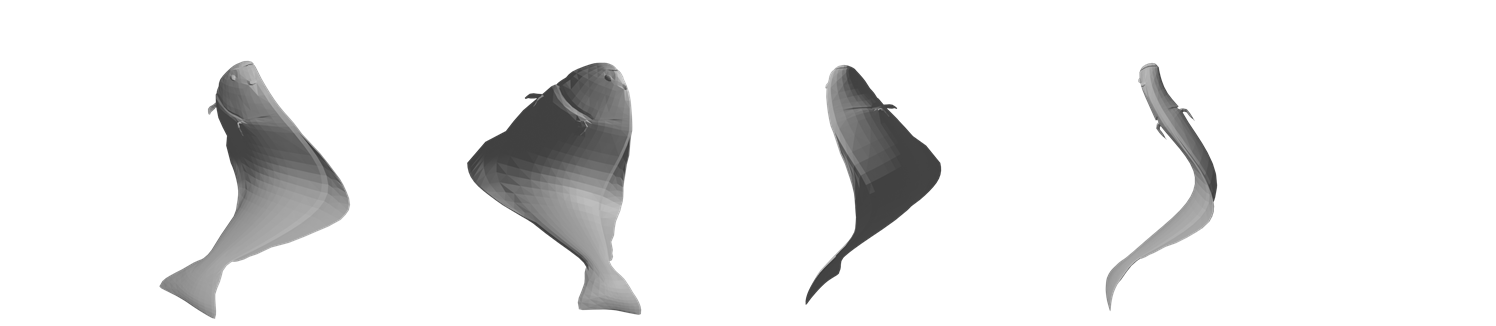}
\caption{Different views of the optimized shape for the fish in Fig.~\ref{pipeline}.}
\label{demo}
\end{figure}

\subsection{Unsupervised Learning}
\label{sec:unsup}
In the deformation model (Sec.~\ref{sec:def}), there are learnable skinning weights, joints' deformation, and root body deformation. In summary, all parameters needed to be optimized are $Q_{j,t}\in R^{3\times3}$, $R_{j,t}\in R^{3\times3}$, $T_{j,t}\in R^3$, $S_{j,t}\in R^1$, $R_{0,t}\in R^{3\times3}$, $T_{0,t}\in R^3$, and $S_{0,t}\in R^1$. For each frame, all these parameters will be optimized in an unsupervised iterative learning manner based on some 2D some reconstruction losses, without the use of any 3D shape ground-truth nor multi-view images, by leveraging the power of differentiable renders~\cite{liu2019softras, ravi2020pytorch3d}.

\textbf{Forward Rendering} Given a 3D deformed mesh template from our deformation model, by using a differentiable render~\cite{liu2019softras, ravi2020pytorch3d}, which consists of a mesh rasterizer and a silhouette shader, we can obtain its 2D projected silhouette, denoted as $M_{template}$, which can be compared with the pseudo-ground-truth silhouette of the cropped target fish, $M_{target}$, obtained from a pre-trained U-Net~\cite{ronneberger2015u} for instance segmentation, as shown in Fig.~\ref{pipeline}.

\textbf{Reconstruction Losses} We iteratively minimize the following 2D reconstruction losses between $M_{template}$ and $M_{target}$ by applying gradient descent on all learnable parameters introduced in Sec.~\ref{sec:def}:
\begin{equation}
\begin{aligned}
{L_{rec}} &=  L_{iou} + L_{boundary} + \lambda_{s} \cdot L_{s} + \lambda_{t} \cdot L_{t} \\ & + \lambda_{n} \cdot L_{normal} + \lambda_{l} \cdot L_{lap}, \\ L_{iou}&=1- \frac{|M_{template} \cap M_{target}|}{|M_{template} \cup M_{target}|}
\end{aligned}
\end{equation}
where $L_{rec}$ is the total reconstruction loss, $L_{iou}$ is one minus the intersection of union (IOU) between $M_{target}$ and $M_{template}$.

IOU is a widely used metric for comparing the whole silhouettes between two segmentation masks but cannot fully reflect the reconstruction performance on the boundary of the silhouette, which normally contains richer information about the 3D shape. Therefore, we further introduce boundary loss, $L_{boundary}$, proposed in a medical image segmentation work~\cite{kervadec2019boundary}, to help the model reconstruct 3D shape based on the 2D silhouette. $L_{boundary}$ is differentiable and implemented based on the distance transform maps of the silhouette images. More details can be found in the work~\cite{kervadec2019boundary}.

As mentioned in Sec.~\ref{sec:def}, joints' scaling and translation are used to minimize the individual fine-grained difference with the shared template, while the coarse-grained deformation is mostly modeled by joints' rotation. Therefore, we add the following scaling regularization, $L_{s}$, and translation regularization, $L_t$ in $L_{rec}$:

\begin{equation}
\begin{aligned}
L_{s} &= \|S_{j,t}-1\|_{2}^{2}, \\ L_{t} &= \|T_{j,t}\|_{2}^{2}.
\end{aligned}
\end{equation}

Additionally, we enforce smoothness of the deformed 3D shape by adding the following two losses, i.e., $L_{normal}$, which enforces consistency across the normal vector of neighboring faces, and $L_{lap}$, which is a Laplacian regularizer on neighboring vertices:

\begin{equation}
\begin{aligned}
L_{normal} &= \sum_{i,j} {(1- cos(n_{i,t}, n_{j,t}))}, \\ L_{lap} &=\left\|{{V}}_{i,t}-\frac{1}{\left|N_{i}\right|} \sum_{j \in N_{i}} {{V}}_{j,t}\right\|_{2}^{2},
\end{aligned}
\end{equation}
where $n_{i,t}$ is the normal vector of one face on the mesh and $n_{j,t}$ is the normal vector of one neighboring face; $cos$ is the cosine similarity. $V_{i,t}$ is one mesh vertex location and $V_{j,t}$ is one of its neighboring mesh vertices.

More specifically, $\lambda_{s}$ is 1, $\lambda_{t}$ is 10, $\lambda_{n}$ and $\lambda_{l}$ are 0.003. With these loss functions calculated only based on 2D silhouettes, the model can iteratively learn the 3D shape without the use of any 3D shape ground-truth or multi-view images. Fig.~\ref{demo} shows the different views of the optimized shape for the fish in Fig.~\ref{pipeline}. As shown in Fig.~\ref{pipeline}, three pre-defined keypoints, i.e., head, center, and tail (red, blue, and green) keypoints on the final optimized shape, $S_t$, will be delivered to 3D keypoints localization module for the downstream task, length measurement.  

\subsection{Absolute 3D Localization}
\label{sec:Loc}
\begin{figure}[htpb]
\centering
\includegraphics[width=0.5\textwidth]{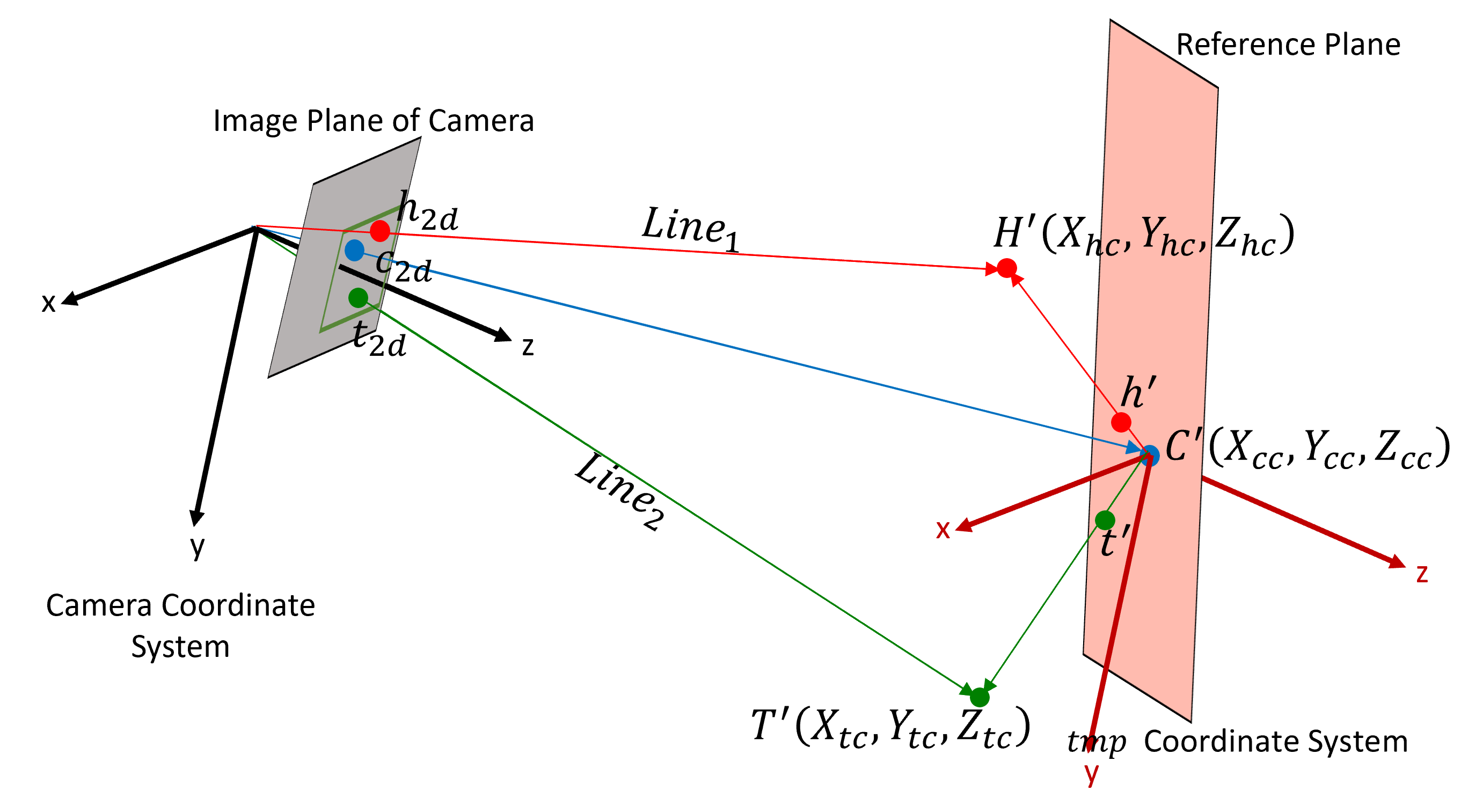}
\caption{3D keypoints localization: $h'$, $C'$, and $t'$ define the relative 3D pose. $H'$, $C'$, and $T'$ defines the absolute 3D pose.}
\label{3D localization}
\end{figure}

The differentiable render in Sec.~\ref{sec:unsup} generates the silhouette of $S_t$ as well as the projection of three keypoints. Our 3D keypoints localization module also takes the location of three pre-defined keypoints in 2D image as input.  Fig.~\ref{pipeline} shows these three keypoints' 2D locations in the output frame. We denote these three keypoints on $S_t$ as $h$ $(x_h, y_h, z_h)$, $c$ $(x_c, y_c, z_c)$, and $t$ $(x_t, y_t, z_t)$ respectively. Their 2D locations in the frame are denoted as $h_{2d}$ $(U_h,V_h)$, $c_{2d}$ $(U_c,V_c)$, $t_{2d}$ $(U_t,V_t)$ respectively in the image coordinate system, as shown in Fig.~\ref{3D localization}.

The 3D keypoints localization module aims to locate three keypoints in absolute 3D of world coordinate system, as denoted in Fig.~\ref{3D localization} respectively as $H'$ $(X_{hc}, Y_{hc}, Z_{hc})$, $C'$ $(X_{cc}, Y_{cc}, Z_{cc})$, and $T'$ $(X_{tc}, Y_{tc}, Z_{tc})$, where the second subscript $c$ denotes the camera coordinate system and the unit is $millimeter$. Following~\cite{mei2021absolute}, this task is formulated as the following closed-form solution. 

\textbf{Back Projection} Using camera intrinsic parameters, $K\in R^{3\times3}$, we can back project $h_{2d}$, $c_{2d}$, $t_{2d}$ to 3D space in the camera coordinate system:
\begin{equation}
\left[
\begin{array}{c}
X \\ Y \\ 1
\end{array}\right]
=K^{-1}\left[\begin{array}{c}
U \\ V \\ 1
\end{array}\right].
\end{equation}
 Now for thee keypoints, respectively we get $h''$ $(X_h, Y_h, 1)$, $c''$ $(X_c, Y_c, 1)$, and $t''$ $(X_t, Y_t, 1)$ in the camera coordinate system without depth. But the head point $H'$ and tail point $T'$ in the absolute 3D in the camera coordinate system must be on the following two lines respectively, as shown in Fig.~\ref{3D localization}:
 
\begin{equation}
\begin{aligned}
Line_1: (X, Y, Z)&=m \cdot (X_{h}, Y_{h},1), m \in R, \\ Line_2: (X, Y, Z)&=n \cdot (X_{t}, Y_{t},1), n \in R.
\end{aligned}
\label{2 lines}
\end{equation}

Their exact locations can thus be calculated with the help of the following reference plane.

\textbf{Reference Plane} As pointed out in the longline fishing works~\cite{huang2018fish, williams2016automated}, all hooked fish are pulled up from the water by a fishing rail line. Therefore, our assumption follows~\cite{mei2021absolute} that all fish's center points are always on a reference plane shown in Fig.~\ref{3D localization}, which coincides with the plane of a checkerboard of known grid size used in the camera calibration, defined as $Z = 0$ plane in the world coordinate system. This world coordinate system is illustrated in the right top corner of Fig.\ref{pipeline}. 

Using solvePnP method, we can calculate the rotation matrix, $R_{3\times3}$, and translation vector, $T_{3\times1}$, between the world coordinate system and the camera coordinate system. Therefore, the homography matrix $H$ between the image plane and $Z=0$ plane in the world coordinate system is obtained by Eq.\ref{H}:
\begin{equation}
\begin{aligned}
H \quad = \quad K_{3\times3} & \cdot\left[\begin{array}{lll}
R_{1} & R_{2} & T
\end{array}\right]_{3\times3},
\label{H}
\end{aligned}
\end{equation}
where $R_{1},\ R_{2}$ are the first two columns in $R_{3\times3}$.

Under our assumption, center point $C'$ is a point $(X_{cw}, Y_{cw}, 0)$ in the world coordinate system. As a result, with homography matrix $H$, we can get $Z_{cc}$ (depth) with Eq.\ref{Zcc}

\begin{equation}
\begin{aligned}
\left[\begin{array}{c}
X_{cw}/ Z_{cc} \\ Y_{cw}/ Z_{cc} \\ 1/Z_{cc}
\end{array}\right]&=H^{-1} \cdot\left[\begin{array}{c}
U_{c} \\ V_{c} \\ 1
\end{array}\right],
\label{Zcc}
\end{aligned}
\end{equation}

Finally, with $Z_{cc}$ (depth), we can get the center point $C'$ in the absolute 3D in the camera coordinate system, as shown in Fig.~\ref{3D localization}:
\begin{equation}
\label{eq:center}
(X_{cc}, Y_{cc}, Z_{cc}) = Z_{cc}\cdot (X_c, Y_c, 1).
\end{equation}

\textbf{3D Localization} Now we want to estimate the absolute fish pose centered at $C'$. In order to locate $H'$ on $Line_1$ and $T'$ on $Line_2$, as illustrated in Fig.\ref{3D localization}, we set up a new coordinate system, called $tmp$, whose origin point is at $C'$ and has no relative rotation with respect to the camera coordinate system, while there is a translation vector between them being $(X_{cc}, Y_{cc}, Z_{cc})$. Then we translate $h$, $c$ and $t$ to the $tmp$ coordinate system by translating $c$ to $tmp$'s origin point:
\begin{equation}
\begin{aligned}
c'&=(X_{cc}, Y_{cc}, Z_{cc}), \\ h'&=(x_h-x_c, y_h-y_c, z_h-z_c) +  (X_{cc}, Y_{cc}, Z_{cc}), \\ t'&=(x_t-x_c, y_t-y_c, z_t-z_c) +  (X_{cc}, Y_{cc}, Z_{cc}),
\end{aligned}
\end{equation}
where $t'$ $(x_{tc}, y_{tc}, z_{tc})$ and $h'$ $(x_{hc}, y_{hc}, z_{hc})$ are relative head point and tail point in the camera coordinate system. Note that the absolute head point, $H'$, and tail point, $T'$, must also lie on the following lines $C'h'$ and $C't'$ respectively:
\begin{small}
\begin{equation}
\begin{aligned}
(X, Y, Z)_h&=a \cdot (x_{hc}, y_{hc}, z_{hc}) + (1-a) \cdot (X_{cc}, Y_{cc}, Z_{cc}), a \in R, \\ (X, Y, Z)_t&=b \cdot (x_{tc}, y_{tc}, z_{tc}) + (1-b) \cdot (X_{cc}, Y_{cc}, Z_{cc}), b \in R.
\end{aligned}
\end{equation}
\end{small}

As a result, the intersection point of line $C'h'$ and $Line_1$ is $H'$, and $T'$ is the intersection point of line $C't'$ and $Line_2$. Same as~\cite{mei2021absolute}, we use the simple least-squares method to calculate the intersection points. Finally, the absolute fish length can thus be calculated:
\begin{equation}
\begin{aligned}
{Length} &= {\wideparen{H'C’T'}}= {\mid H'T'\mid}\cdot{\frac{\overset{\frown} {h' t'}}{\mid h't'\mid}},
\end{aligned}
\end{equation}
where $\frac{\overset{\frown} {h' t'}}{\mid h't'\mid}$ is the bending ratio calculated based on optimized shape $S_t$. The right bottom corner in Fig.\ref{pipeline} is a 3D localization demo.

\begin{table}[htbp]
    \caption{Comparison Evaluation and Ablation Study}\label{evaluation} 
    \setlength{\tabcolsep}{1mm}{\begin{tabular}{cccccc}
        \toprule[1.5pt]
        Method & Bias(mm) & EMD(mm) & RMSD & KL \\
        \midrule[0.5pt]
        Stereo~\cite{huang2018fish} & -40.5 & 46.0 & 7.9\% & 0.26  \\
        MonoPose~\cite{mei2021absolute} & -9.3 & 43.1 & 7.3\% &0.23  \\
        BFS & -10.2 & 24.2 & 5.6\% & 0.11  \\
        MonoPose w/o Bending & -95.4 & 99.3 & 10.4\% &0.53  \\
        BFS w/o Bending & -55.4 & 60.0 & 7.9\% & 0.28  \\
        Ours w/o Bending & -48.8 & 62.1 & 8.7\% &0.36  \\
        \midrule[0.5pt]
        Ours (DMR) & 4.5 & 37.1 & 7.1\% &0.21  \\
        \bottomrule[1.5pt]
    \end{tabular}}
\end{table}

\section{Experiments}
\textbf{Dataset \& Evaluation} Following~\cite{huang2018fish, mei2021absolute}, we use a fishing dataset containing only the normalized lengths histogram (distribution) of 738 ground-truth length labeled fish and the corresponding several hours' fish-catching stereo videos, without one-to-one correspondence of the ground-truth length of each individual captured fish. Video input allows our model to average the estimated length for all frames corresponding to each individual fish. To fairly compare with previous works, we also use the difference between predicted length histogram and ground truth length histogram for evaluation. Specifically, the metrics are bias, root mean square deviation (RMSD), Kullback-Leibler (KL) divergence~\cite{kullback1951information}, and earth mover's distance (EMD)~\cite{levina2001earth}. For these four metrics, lower absolute value means better histogram estimation.

\begin{figure}[htpb]
\centering
\includegraphics[width=0.4\textwidth]{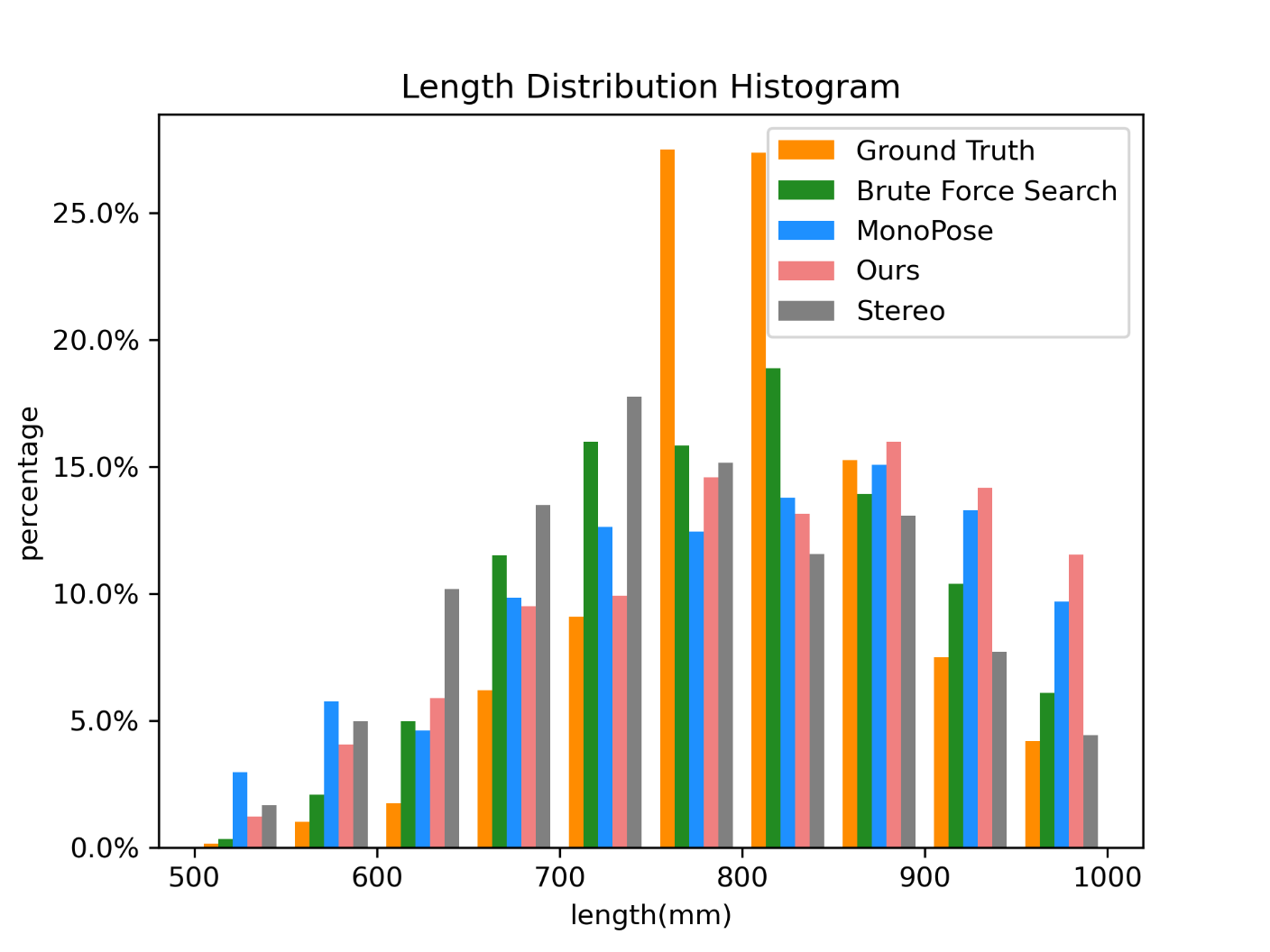}
\caption{Length histogram (distribution): Only lengths in $[500, 1000]mm$ are kept because most lengths are within this range. $y$ axis is percentage of fish number in bins over total number. $x$ axis is length bins.}
\label{comparsion}
\end{figure}

\textbf{Results}
The stereo method~\cite{huang2018fish} requires stereo image pairs as input and is more difficult to deploy in the challenging at-sea environment, on the other hand, ours only needs a single-view image. Although MonoPose~\cite{mei2021absolute} also only requires a single-view image, it uses a flat surface with simple deformation to model the shape. Fig.\ref{comparsion} and Table.\ref{evaluation} show superior performance of our method in all four metrics.

\textbf{Ablation Study} We also report the performance of brute force search (BFS), which exhaustively searches for all possible (millions) deformation combinations, from work~\cite{mei2021absolute}. Fig.\ref{comparsion} and Table.\ref{evaluation} show our unsupervised learning method achieves favorable performance as BFS, which is much more memory demanding, and its performance is limited to the database size~\cite{mei2021absolute}. Besides, we remove bending ratio for BFS, MonoPose~\cite{mei2021absolute}, and ours. Table.\ref{evaluation} shows severely deformation modeling is critical in this downstream task.

\section{Conclusions}
We present DMR, a monocular unsupervised learning method for severely deformed shape reconstruction based on a 3D deformable template. DMR can faithfully reconstruct severely deformed objects like fish and be applied to downstream tasks which require consistent semantic meaning of mesh vertices. The evaluation shows DMR's superior performance.

\bibliographystyle{IEEEbib}
\bibliography{icme2022template}

\end{document}